# Bayesian Multicategory Support Vector Machines

**Zhihua Zhang**
Electrical and Computer Engineering
University of California
Santa Barbara, CA 93106

**Michael I. Jordan**
Computer Science and Statistics
University of California
Berkeley, CA 94720

## Abstract

We show that the multi-class support vector machine (MSVM) proposed by Lee et al. (2004) can be viewed as a MAP estimation procedure under an appropriate probabilistic interpretation of the classifier. We also show that this interpretation can be extended to a hierarchical Bayesian architecture and to a fully-Bayesian inference procedure for multi-class classification based on data augmentation. We present empirical results that show that the advantages of the Bayesian formalism are obtained without a loss in classification accuracy.

## 1 Introduction

The support vector machine (SVM) is a popular classification methodology (Vapnik, 1998; Cristianini and Shawe-Taylor, 2000). While widely deployed in practical problems, two issues limit its practical applicability—its focus on binary classification and its inability to provide estimates of uncertainty. Methods for handling the multi-class problem have slowly emerged. In parallel, probabilistic approaches to large-margin classification have begun to emerge. In this paper we provide a treatment that combines these themes—we provide a Bayesian treatment of the multi-class problem.

A variety of adhoc methods for extending the binary SVM to multi-class problems have been studied; these include one-versus-all, one-versus-one, error-correcting codes and pairwise coupling (Allwein et al., 2000). One of the more principled approaches to the problem is due to Lee et al. (2004). These authors proposed a multi-class SVM (MSVM) which treats the multiple classes jointly. They prove that their MSVM satisfies a Fisher consistency condition, a desirable property that does not hold for many other multi-class SVMs (see, e.g., Vapnik, 1998; Bredensteiner and Bennett, 1999; Weston and Watkins, 1999; Crammer and Singer, 2001; Guermeur, 2002).

In this paper we show that the MSVM framework of Lee et al. (2004) also has the advantage that it can be extended to a Bayesian model. We show how this can be achieved in two stages. First, we concern ourselves with interpreting the cost function underlying the MSVM as a likelihood and show that the MSVM estimation procedure can be viewed as a MAP estimation procedure under an appropriate prior. Second, we introduce a latent variable representation of the likelihood function and show how this yields a full Bayesian hierarchy to which data augmentation algorithms can be applied.

Our work is related to earlier papers by Sollich (2001) and Mallick et al. (2005), who presented probabilistic interpretations of binary SVMs for classification, and Chakraborty et al. (2005) who investigated a fully Bayesian support vector regression method. Our approach borrows some of the technical machinery from these papers in the construction of likelihood functions, but our focus on the MSVM framework leads us in a somewhat different direction from those papers, none of which readily yield multi-class SVMs.

The major advantage of the Bayesian approach is that it provides an estimate of uncertainty in its predictions—an important desideratum in real-world applications that is missing in both the SVM and the MSVM. It is important, however, to assess whether this gain is achieved at the expense of a loss in classification accuracy. Indeed, the fact that the SVM and the MSVM are based on cost functions that are surrogates of classification error makes these methods state-of-the-art in terms of classification accuracy. We investigate this issue empirically in this paper.

The rest of this paper is organized as follows. Section 2 reviews the basic principles of the MSVM and shows how it can be formulated as a MAP estimation pro-cedure. Section 3 extends this formulation to a fully Bayesian hierarchical model. Sections 4 presents an inferential methodology for this model. Experimental results are presented in Section 5, and concluding remarks are given in Section 6.

## 2 Probabilistic Multicategory Support Vector Machines

Consider a classification problem with $c$ classes. We are given a set of training data $\{\mathbf{x}_i, y_i\}_1^n$ where $\mathbf{x}_i \in \mathbb{R}^p$ is an input vector and $y_i$ represents its class label. We let $y_i$ be a multinomial variable indicating class membership, i.e., $y_i \in \{1, \ldots, c\}$, where $y_i = j$ indicates that $\mathbf{x}_i$ belongs to class $j$. We let $\mathbf{1}_m$ denote the $m \times 1$ vector of 1's, let $\mathbf{I}_m$ denote the $m \times m$ identity matrix, and let $\mathbf{0}$ denote the zero vector (or matrix) whose dimensionality is dependent upon the context. In addition, $\mathbf{A} \otimes \mathbf{B}$ represents the Kronecker product of $\mathbf{A}$ and $\mathbf{B}$.

### 2.1 Multicategory Support Vector Machines

The MSVM (Lee et al., 2004) is based on a $c$-tuple of classification functions, $\mathbf{f}(\mathbf{x}) = (f_1(\mathbf{x}), \ldots, f_c(\mathbf{x}))$, which respect the constraint $\sum_{j=1}^c f_j(\mathbf{x}) = 0$, for any $\mathbf{x} \in \mathbb{R}^p$. Each $f_j(\mathbf{x})$ is assumed to take the form $h_j(\mathbf{x}) + b_j$ with $h_j(\mathbf{x}) \in \mathcal{H}_K$, where $\mathcal{H}_K$ is an RKHS. That is, $\mathbf{f}(\mathbf{x}) = (f_1(\mathbf{x}), \ldots, f_c(\mathbf{x})) \in \prod_{j=1}^c (\{1\} + \mathcal{H}_K)$. The problem of estimating the parameters of the MSVM is formulated as a regularization problem:

$$\min_{\mathbf{f}} \left\{ \sum_{i=1}^n \sum_{j \neq y_i} \left( f_j(\mathbf{x}_i) + \frac{1}{c-1} \right)_+ + \frac{\gamma}{2} \sum_{j=1}^c \|h_j\|_K \right\},$$

where $(u)_+ = u$ if $u > 0$ and $(u)_+ = 0$ otherwise, $\|\cdot\|_K$ is the RKHS norm and $\gamma > 0$ is the regularization parameter. Note that the first term can be interpreted as a data misfit term under an encoding in which the label is 1 in the $j$th element and $-1/(c-1)$ elsewhere. Lee et al. (2004) show that the representer theorem (Kimeldorf and Wahba, 1971) holds for this optimization problem; the optimizing $f_j(\mathbf{x})$ necessarily takes the form

$$f_j(\mathbf{x}) = w_{0j} + \sum_{i=1}^n w_{ij} K(\mathbf{x}, \mathbf{x}_i), \quad (1)$$

under the constraint $\sum_{j=1}^c f_j(\mathbf{x}_i) = 0$ for $i = 1, \ldots, n$. Here $K(\cdot, \cdot)$ is the reproducing kernel (Aronszajn, 1950).

Based on the representer theorem, the MSVM can be reformulated as the following primal optimization problem:

$$\min_{\mathbf{w}_0, \mathbf{W}} \left\{ \sum_{i=1}^n \sum_{j \neq y_i} \left( f_j(\mathbf{x}_i) + \frac{1}{c-1} \right)_+ + \frac{\gamma}{2} \text{tr}(\mathbf{KWW}') \right\}$$
(2)

subject to
$$(\mathbf{w}_0' \mathbf{1}_c) \mathbf{1}_n + \mathbf{KW} \mathbf{1}_c = \mathbf{0}, \quad (3)$$

where $\mathbf{K} = [K(\mathbf{x}_i, \mathbf{x}_j)]$ is the $n \times n$ kernel matrix, $\mathbf{w}_0 = (w_{01}, \ldots, w_{0c})'$ is the $c \times 1$ vector of intercept terms, and $\mathbf{W} = [w_{ij}]$ is the $n \times c$ matrix of regression coefficients. For a new input vector $\mathbf{x}_*$, the classification rule induced by $\mathbf{f}(\mathbf{x}_*)$ is to label $y_* = \arg\max_j f_j(\mathbf{x}_*)$.

### 2.2 Probabilistic Formulation

It is easily seen that a sufficient condition for (3) is

$$\mathbf{w}_0' \mathbf{1}_c = 0 \quad \text{and} \quad \mathbf{W} \mathbf{1}_c = \mathbf{0}. \quad (4)$$

This suggests the following reparameterization:

$$\mathbf{w}_0 = \mathbf{H} \mathbf{b}_0 \quad \text{and} \quad \mathbf{W} = \mathbf{BH}, \quad (5)$$

where $\mathbf{H} = \mathbf{I}_c - \frac{1}{c} \mathbf{1}_c \mathbf{1}_c'$ is the centering matrix, $\mathbf{b}_0$ is an $c \times 1$ vector and $\mathbf{B}$ is an $n \times c$ matrix. Given that $\mathbf{H} \mathbf{1}_c = \mathbf{0}$, the conditions (4) are naturally satisfied, and thus so are the conditions in (3). Thus we parameterize the model in terms of $\mathbf{b}_0$ and $\mathbf{B}$ rather than $\mathbf{w}_0$ and $\mathbf{W}$. Note that (3) is required to hold for any $n \times n$ kernel matrix over the input space; thus, although suggestion (5) is not necessary for (3), it is a weak sufficient condition.

To develop a probabilistic approach to the MSVM, we need to develop a conditional probabilistic model that yields a likelihood akin to the cost used in the MSVM. Our starting point is the following assumption:

$$p(y_i = j | \mathbf{f}(\mathbf{x}_i)) \propto \exp \left\{ - \sum_{l \neq j} \left( f_l(\mathbf{x}_i) + \frac{1}{c-1} \right)_+ \right\}. \quad (6)$$

We also assume that the $y_i$ are conditionally independent given the $\mathbf{f}(\mathbf{x}_i)$, $i = 1, \ldots, n$. Given these assumptions, the unnormalized joint conditional probability of the class labels given the covariate vectors takes the following form:

$$p\left(\{y_i\}_{i=1}^n | \{\mathbf{f}(\mathbf{x}_i)\}_{i=1}^n\right)$$
$$\propto \prod_{i=1}^n \exp \left\{ - \sum_{j \neq y_i} \left( f_j(\mathbf{x}_i) + \frac{1}{c-1} \right)_+ \right\} \quad (7)$$
$$= \prod_{j=1}^c \exp \left\{ - \sum_{\mathbf{x}_i \notin \mathcal{X}_j} \left( f_j(\mathbf{x}_i) + \frac{1}{c-1} \right)_+ \right\}, \quad (8)$$

where $\mathcal{X}_j$ represents the set of input vectors belonging to the $j$th class.

The normalizing constant of the likelihood may involve **B**, and thus we write the likelihood as follows:

$$p(\mathbf{y}|\mathbf{B}) = \frac{1}{g(\mathbf{B})} \prod_{i=1}^{n} \exp\left\{-\sum_{\mathbf{x}_i \notin \mathcal{X}_j} \left(f_j(\mathbf{x}_i) + \frac{1}{c-1}\right)_+\right\},$$

where $\mathbf{y} = (y_1,\ldots,y_n)$ and $g(\mathbf{B})$ is the normalizing constant.

To deal with the normalizing constant, we make use of a multi-class extension of a method proposed by Sollich (2001) in the binary setting. We first rewrite (6) as $p(y_i{=}j|\mathbf{f}(\mathbf{x}_i)) = \prod_{l=1, l\neq j}^{c} p(y_i{\neq}l|f_l(\mathbf{x}_i))$ with

$$p(y_i{\neq}l|f_l(\mathbf{x}_i)) = \frac{1}{g_l(\mathbf{b}_{\cdot l})} \exp\left\{-\left(f_l(\mathbf{x}_i) + \frac{1}{c-1}\right)_+\right\} \quad (9)$$

where $\mathbf{b}_{\cdot l}$ is the $l$th column of **B**. Note that we here make the assumption that the $\mathbf{b}_{\cdot l}$ are mutually independent. Hence, we can express $g(\mathbf{B}) = \prod_{l=1}^{c} g_l(\mathbf{b}_{\cdot l})$, and so (9). We then define $p(\mathbf{b}_{\cdot l}) \propto q(\mathbf{b}_{\cdot j}) g_l(\mathbf{b}_{\cdot l})$ to eliminate the normalizing constants $g_l(\mathbf{b}_{\cdot l})$. Letting $q(\mathbf{b}_{\cdot l}) = \mathcal{N}(\mathbf{0}, \lambda^{-1}\mathbf{K}^{-1})$ and again making use of (8), we obtain the following expression for the joint distribution of the data and the parameter:

$$p(\mathbf{y}, \mathbf{B}) \propto \prod_{j=1}^{c} \exp\left\{-\sum_{\mathbf{x}_i \notin \mathcal{X}_j} \left(f_j(\mathbf{x}_i) + \frac{1}{c-1}\right)_+\right\} q(\mathbf{b}_{\cdot j}).$$

If desired, a MAP estimate of **B** can be obtained by maximizing (the logarithm of) this expression with respect to **B**.

It is well known that the kernel matrix **K** derived from the Gaussian kernel function is positive definite and thus nonsingular. For other kernels, however, we may obtain a singular kernel matrix **K**. For such a **K**, we use its Moore-Penrose inverse $\mathbf{K}^+$ instead, in which case the prior distribution of $\mathbf{b}_{\cdot l}$ becomes a singular normal distribution (Mardia et al., 1979). In either case, we use the notation $\mathbf{K}^{-1}$ for simplicity.

The assumption that $q(\mathbf{B}) = \prod_{j=1}^{c} \mathcal{N}(\mathbf{b}_{\cdot j}|\mathbf{0}, \lambda^{-1}\mathbf{K}^{-1})$ implies that $q(\mathbf{B})$ is a matrix-variate normal distribution with mean matrix **0** and covariance matrix $\lambda^{-1}\mathbf{K}^{-1} \otimes \mathbf{I}_c$, denoted $\mathcal{N}_{n,c}(\mathbf{0}, \lambda^{-1}\mathbf{K}^{-1}\otimes\mathbf{I}_c)$ (Gupta and Nagar, 2000). Given the relationship $\mathbf{W} = \mathbf{BH}$, it turns out that $q(\mathbf{W}) = \mathcal{N}_{n,c}(\mathbf{W}|\mathbf{0}, \lambda^{-1}\mathbf{K}^{-1}\otimes\mathbf{H})$, is a singular matrix-variate normal distribution with the following density function:

$$q(\mathbf{W}) \propto \lambda^{\frac{n(c-1)}{2}} |\mathbf{K}|^{\frac{c-1}{2}} \exp\left\{-\frac{\lambda}{2}\text{tr}(\mathbf{KWW}')\right\}. \quad (10)$$

The derivation of this result is given in the Appendix. It readily follows from (7) and (10) that the MAP estimate of **W** is equivalent to the primal problem for the MSVM where $\lambda$ plays the role of the regularization parameter $\gamma$ in (2).

## 3 Hierarchical Model

In this section we present a hierarchical Bayesian model that completes the likelihood specification presented in the previous section and yields a fully Bayesian approach to the MSVM. We make use of a latent variable representation that makes our conditional independence assumptions explicit and leads naturally to a data augmentation methodology for posterior inference and prediction.

We introduce an $n{\times}c$ matrix $\mathbf{Z} = [z_{ij}]$ of latent variables and impose the constraint $\mathbf{Z}\mathbf{1}_c = \mathbf{0}$. On the one hand, we relate the $z_{ij}$ to the $y_i$ as follows:

$$p(\mathbf{y}|\mathbf{Z}) \propto \prod_{j=1}^{c} \exp\left\{-\sum_{\mathbf{x}_i \notin \mathcal{X}_j} \left(z_{ij} + \frac{1}{c-1}\right)_+\right\}. \quad (11)$$

On the other hand, we need to establish the connection between the $z_{ij}$ and the $f_j(\mathbf{x}_i)$.

Henceforth, we denote $N = \{1,\ldots,n\}$ and $C = \{1,\ldots,c\}$. Assume that $\mathcal{X}_j = \{\mathbf{x}_i : i \in I_j\}$ with $\cup_{j=1}^{c} I_j = N$ and $I_j \cap I_l = \emptyset$ for $j \neq l$. In addition, let $\bar{I}_j$ be the complement of $I_j$ over $N$. Let $n_j$ be the cardinality of $\bar{I}_j$ for $j \in C$. We then have that $\sum_{j=1}^{c} n_j = (c{-}1)n$ due to $\sum_{j=1}^{c}(n{-}n_j) = n$. Recall that $\mathbf{Z}\mathbf{1}_c = 0$, so **Z** has $(c{-}1)n$ degrees of freedom. Moreover, for $j \in C$ and $i \in I_j$, we have

$$z_{ij} = -\sum_{l \neq j} z_{il}. \quad (12)$$

It is clear that $i \in I_j$ if and only if $i \in \bar{I}_l$ with $l \neq j$. This shows that $\mathcal{S} = \{z_{ij} : j \in C, i \in \bar{I}_j\}$ is a linearly independent set. In addition, there only involve $z_{ij} \in \mathcal{S}$ in (11). Thus, we only need to impute the $z_{ij} \in \mathcal{S}$.

Let $\widetilde{\mathbf{K}} = [\mathbf{1}_n, \mathbf{K}]$ and $\widetilde{\mathbf{B}}' = [\mathbf{b}_0, \mathbf{B}']$ be $n{\times}(n{+}1)$ and $c{\times}(n{+}1)$ matrices, respectively. We now associate the $z_{ij} \in \mathcal{S}$ with the $f_j(\mathbf{x}_i)$ via

$$z_{ij} = \tilde{\mathbf{k}}_i'\boldsymbol{\beta}_j + e_{ij} \text{ with } e_{ij} \sim \mathcal{N}(0,\sigma^2), \quad (13)$$

where $\tilde{\mathbf{k}}_i'$ is the $i$th row of $\widetilde{\mathbf{K}}$ and $\boldsymbol{\beta}_j$ $((n{+}1){\times}1)$ is the $j$th column of $\widetilde{\mathbf{B}}$. Denote

$$\mathbf{s}_j = \mathbf{Z}(\bar{I}_j, \{j\}) \text{ and } \widetilde{\mathbf{K}}_j = \widetilde{\mathbf{K}}(\bar{I}_j, :), \ j \in C,$$

where $\mathbf{Z}(\bar{I}_j, \{j\})$ represents the sub-vector of the $j$th column of **Z** containing the elements indexed by $\bar{I}_j$, and $\widetilde{\mathbf{K}}(\bar{I}_j, :)$ is the submatrix of the rows of $\widetilde{\mathbf{K}}$ indexed by $\bar{I}_j$. Thus, $\mathbf{s}_j$ is $n_j{\times}1$ and $\widetilde{\mathbf{K}}_j$ is $n_j{\times}(n{+}1)$. The $\mathbf{s}_j$ are mutually independent. Furthermore, we can express (13) in matrix notation as

$$\mathbf{s}_j = \widetilde{\mathbf{K}}_j\boldsymbol{\beta}_j + \boldsymbol{\epsilon}_j \text{ with } \boldsymbol{\epsilon}_j \sim \mathcal{N}(0,\sigma^2\mathbf{I}_{n_j}) \quad (14)$$

for $j = 1, \ldots, c$. Note that the set containing the elements of $\mathbf{s}_j$, $j = 1, \ldots, c$, is the same as $\mathcal{S}$. We shall interchangeably use $\mathcal{S}$ and $\mathbf{s}_j$'s according to our purpose.

Given the $z_{ij}$, we assume that the labels $y_i$ are independent of $\mathbf{K}$, $\widetilde{\mathbf{B}}$, and $\sigma^2$. Within this conditional independence model, we assign conjugate priors to the parameters $\widetilde{\mathbf{B}}, \sigma^2$ and $\tau$. In particular, we assume that

$$\begin{aligned}\widetilde{\mathbf{B}}, \sigma^2 &\sim \mathcal{G}(\sigma^{-2}|a_\sigma/2, b_\sigma/2)\,\mathcal{N}(\mathbf{b}_0|\mathbf{0}, \sigma^2\eta^{-1}\mathbf{I}_c) \\ &\quad \times \mathcal{N}_{n,c}(\mathbf{B}|\mathbf{0},\, (\tau\mathbf{K})^{-1}\otimes\mathbf{I}_c),\end{aligned} \quad (15)$$

where $\mathcal{G}(u|a, b)$ is a Gamma distribution. Through simple algebraic calculations, we can equivalently express (15) as

$$\widetilde{\mathbf{B}}, \sigma^2 \sim \mathcal{G}\left(\sigma^{-2}\Big|\frac{a_\sigma}{2},\, \frac{b_\sigma}{2}\right)\prod_{j=1}^c\mathcal{N}(\boldsymbol{\beta}_j|\mathbf{0}, \sigma^2\boldsymbol{\Sigma}^{-1}), \quad (16)$$

where $\boldsymbol{\Sigma} = \begin{bmatrix} \eta & \mathbf{0} \\ \mathbf{0} & \tau\mathbf{K} \end{bmatrix}$. Further, we assign

$$\tau \sim \mathcal{G}(\tau|a_\lambda/2,\, b_\lambda/2). \quad (17)$$

In addition, we let the kernel function $K$ be indexed by a parameter $\boldsymbol{\theta}$. In general $\boldsymbol{\theta}$ plays the role of a scale parameter or a location parameter and we endow $\boldsymbol{\theta}$ with the appropriate noninformative prior over $[a_\theta, b_\theta]$ for these roles. In practice, however, we expect that computational issues will often preclude full inference over $\boldsymbol{\theta}$ and in this case we suggest using an empirical Bayes approach in which $\boldsymbol{\theta}$ is set via maximum marginal likelihood.

In summary, we form a hierarchical model, which can be represented as a directed acyclic graph as shown in Figure 1.

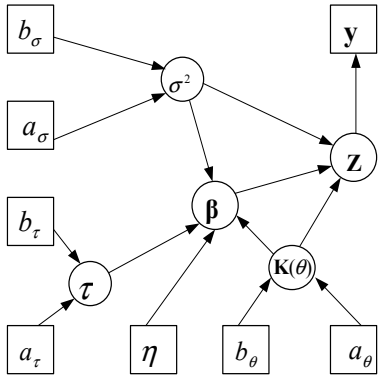

Figure 1: A directed acyclic graph for Bayesian analysis of the MSVM.

## 4 Inference

Our approach to inference for the Bayesian MSVM is based on a data augmentation methodology. Specifically, we base our work on the approach of Albert and Chib (1993), whose work on Bayesian inference in binary and polychotomous regression has been widely used in Bayesian approaches to generalized linear models (Dey et al., 1999).

The overall approach is a special case of the IP algorithm, which consists of two steps: the I-step ("Imputation") and the P-step ("Posterior") (Tanner and Wong, 1987). For our hierarchical model, the I-step first draws a value of each of the variables in $\mathcal{S}$ from the predictive distributions $p(\mathcal{S}|\widetilde{\mathbf{B}}, \boldsymbol{\theta}, \tau, \sigma^2, \mathbf{y})$ ($= p(\mathcal{S}|\widetilde{\mathbf{B}}, \boldsymbol{\theta}, \sigma^2, \mathbf{y})$). The P-step then draws values of the parameters from their complete-data posterior $p(\widetilde{\mathbf{B}}, \boldsymbol{\theta}, \tau, \sigma^2|\mathbf{y}, \mathcal{S})$, which can be simplified to

$$p(\widetilde{\mathbf{B}}, \boldsymbol{\theta}, \tau, \sigma^2|\mathbf{y}, \mathcal{S}) = p(\widetilde{\mathbf{B}}, \boldsymbol{\theta}, \tau, \sigma^2|\mathcal{S}).$$

### 4.1 Algorithm

One sweep of the data augmentation method consists of the following algorithmic steps:

(a) update the latent variables $z_{ij} \in \mathcal{S}$ ;

(b) update the parameters $\boldsymbol{\theta}$, $\widetilde{\mathbf{B}}$ and $\sigma^2$;

(c) update the hyperparameter $\tau$.

Given the initial values of these variables, we run the algorithm for $M1$ sweeps and the first $M2$ sweeps are treated as the burn-in period. After the burn-in, we retain every $M$th realization of the sweep. Hence, we shall keep $T = (M1 - M2)/M$ realizations and use these realizations to predict the class label of test data.

The conditional distribution for the variables $z_{ij} \in \mathcal{S}$ does not have an explicit form, thus, we use a Metropolis-Hastings (MH) sampler to update their values. Given a proposal density $q(\cdot|z_{ij})$, we generate a value $z_{ij}^*$ from $q(z_{ij}^*|z_{ij})$. Note that

$$p(z_{ij}|y_i \neq j, \boldsymbol{\theta}, \widetilde{\mathbf{B}}, \sigma^2) \propto p(y_i \neq j|z_{ij})p(z_{ij}|\boldsymbol{\theta}, \widetilde{\mathbf{B}}, \sigma^2).$$

The acceptance probabilities from the $z_{ij}$ to the $z_{ij}^*$ are then:

$$\beta = \min\left\{1, \frac{p(y_i \neq j|z_{ij}^*)p(z_{ij}^*|\boldsymbol{\theta}, \widetilde{\mathbf{B}}, \sigma^2)q(z_{ij}|z_{ij}^*)}{p(y_i \neq j|z_{ij})p(z_{ij}|\boldsymbol{\theta}, \widetilde{\mathbf{B}}, \sigma^2)q(z_{ij}^*|z_{ij})}\right\},$$

where

$$p(y_i \neq j|z_{ij}) \propto \exp\{-[z_{ij} + 1/(c-1)]_+\}$$

and
$$p(z_{ij}|\boldsymbol{\theta}, \widetilde{\mathbf{B}}, \sigma^2) \propto \exp\{-(z_{ij}-\tilde{\mathbf{k}}'_i\boldsymbol{\beta}_j)^2/(2\sigma^2)\}.$$

In general, the proposal distribution $q(z^*_{ij}|z_{ij})$ is taken as a symmetric distribution (Gilks et al., 1996). For example, we take a normal with mean $z_{ij}$ and a pre-specified standard deviation in the experiments that we report below.

Note that we only need to estimate the $z_{ij} \in \mathcal{S}$; the remaining $z_{ij}$ can be directly calculated from (12). Letting $\mathbf{z}_{\cdot j}$ be the $j$th column of $\mathbf{Z}$, we have from (13) that
$$\mathbf{z}_{\cdot j} = b_{0j}\mathbf{1}_n + \mathbf{K}\mathbf{b}_{\cdot j} + \mathbf{e}_{\cdot j}, \qquad (18)$$
where $\mathbf{e}_{\cdot j}$, $j=1,\ldots,c$, are independent and identically distributed according to $\mathcal{N}(0, \sigma^2\mathbf{I}_n)$. Imposing the constraint $\mathbf{Z}\mathbf{1}_c = \mathbf{0}$, we can further express the above equations in matrix form:
$$\begin{aligned}\mathbf{Z} &= \mathbf{1}_n\mathbf{b}'_0\mathbf{H} + \mathbf{K}\mathbf{B}\mathbf{H} + \mathbf{E}\mathbf{H}, \\ &= \mathbf{1}_n\mathbf{w}_0 + \mathbf{K}\mathbf{W} + \mathbf{E}\mathbf{H},\end{aligned}$$
where $\mathbf{E} = [\mathbf{e}_{\cdot 1},\ldots,\mathbf{e}_{\cdot c}]$ is the $n{\times}c$ error matrix.

Steps (b) and (c) are based on the fact that
$$\begin{aligned}&p(\widetilde{\mathbf{B}}, \boldsymbol{\theta}, \tau, \sigma^2|\mathcal{S}) \\ &\propto p(\widetilde{\mathbf{B}}, \boldsymbol{\theta}, \tau, \sigma^2, \mathcal{S}) = p(\widetilde{\mathbf{B}}, \boldsymbol{\theta}, \tau, \sigma^2, \{\mathbf{s}_j\}_{j=1}^c) \\ &= p(\widetilde{\mathbf{B}}, \boldsymbol{\theta}, \tau, \sigma^2)p(\{\mathbf{s}_j\}_{j=1}^c|\widetilde{\mathbf{B}}, \boldsymbol{\theta}, \sigma^2).\end{aligned}$$

If we wish to include an update $\boldsymbol{\theta}$ in our sampling procedure, then we again need to avail ourselves of a MH sampler. We write the marginal distribution of $\boldsymbol{\theta}$ conditional on the $\mathbf{s}_j$ and $\boldsymbol{\beta}_j$ as
$$p(\boldsymbol{\theta}|\{\mathbf{s}_j, \boldsymbol{\beta}_j\}_{j=1}^c) \propto \left\{\prod_{j=1}^c p(\boldsymbol{\beta}_j|\boldsymbol{\theta})p(\mathbf{s}_j|\boldsymbol{\beta}_j, \boldsymbol{\theta})\right\}p(\boldsymbol{\theta}).$$

Let $\boldsymbol{\theta}^*$ denote the proposed move to the current $\boldsymbol{\theta}$. Then this move is accepted in probability
$$\min\left\{1, \prod_{j=1}^c \frac{p(\boldsymbol{\beta}_j|\boldsymbol{\theta}^*)p(\mathbf{s}_j|\boldsymbol{\beta}_j, \boldsymbol{\theta}^*)}{p(\boldsymbol{\beta}_j|\boldsymbol{\theta})p(\mathbf{s}_j|\boldsymbol{\beta}_j, \boldsymbol{\theta})}\right\}.$$

It follows from (14) that the distribution of $\mathbf{s}_j$ conditional on $\boldsymbol{\beta}_j$ and $\boldsymbol{\theta}$ is a generalized multivariate $t$ distribution (Arellano-Valle and Bolfarine, 1995). That is, for $j \in C$, the p.d.f. is given by
$$p(\mathbf{s}_j|\boldsymbol{\beta}_j, \boldsymbol{\theta}) \propto \left(b_\sigma + (\mathbf{s}_j{-}\widetilde{\mathbf{K}}_j\boldsymbol{\beta}_j)'(\mathbf{s}_j{-}\widetilde{\mathbf{K}}_j\boldsymbol{\beta}_j)\right)^{-\frac{a_\sigma+n_j}{2}}.$$

This yields
$$\frac{p(\mathbf{s}_j|\boldsymbol{\beta}_j, \boldsymbol{\theta}^*)}{p(\mathbf{s}_j|\boldsymbol{\beta}_j, \boldsymbol{\theta})} = \left(\frac{b_\sigma + (\mathbf{s}_j{-}\widetilde{\mathbf{K}}_j\boldsymbol{\beta}_j)'(\mathbf{s}_j{-}\widetilde{\mathbf{K}}_j\boldsymbol{\beta}_j)}{b_\sigma + (\mathbf{s}_j{-}\widetilde{\mathbf{K}}^*_j\boldsymbol{\beta}_j)'(\mathbf{s}_j{-}\widetilde{\mathbf{K}}^*_j\boldsymbol{\beta}_j)}\right)^{\frac{a_\sigma+n_j}{2}},$$

where $\widetilde{\mathbf{K}}^*_j$ is obtained from $\widetilde{\mathbf{K}}_j$ with $\boldsymbol{\theta}^*$ replacing $\boldsymbol{\theta}$. Also, note that the marginal distribution of $\boldsymbol{\beta}_j$ conditional on $\boldsymbol{\theta}$ is a generalized multivariate $t$ distribution
$$p(\boldsymbol{\beta}_j|\boldsymbol{\theta}) \propto |\boldsymbol{\Sigma}|^{\frac{1}{2}}\left(b_\sigma + \boldsymbol{\beta}'_j\boldsymbol{\Sigma}\boldsymbol{\beta}_j\right)^{-\frac{a_\sigma+n+1}{2}}.$$

Given that $|\boldsymbol{\Sigma}| = \eta\tau^n|\mathbf{K}|$, we have
$$\frac{p(\boldsymbol{\beta}_j|\boldsymbol{\theta}^*)}{p(\boldsymbol{\beta}_j|\boldsymbol{\theta})} = \frac{|\mathbf{K}^*|^{\frac{1}{2}}}{|\mathbf{K}|^{\frac{1}{2}}}\left(\frac{b_\sigma + \boldsymbol{\beta}'_j\boldsymbol{\Sigma}\boldsymbol{\beta}_j}{b_\sigma + \boldsymbol{\beta}'_j\boldsymbol{\Sigma}^*\boldsymbol{\beta}_j}\right)^{\frac{a_\sigma+n+1}{2}},$$
where $\mathbf{K}^*$ and $\boldsymbol{\Sigma}^*$ are obtained from $\mathbf{K}$ and $\boldsymbol{\Sigma}$, respectively, with $\boldsymbol{\theta}^*$ replacing $\boldsymbol{\theta}$.

Given the joint prior of $\widetilde{\mathbf{B}}$ and $\sigma^{-2}$ in (16), their joint posterior density conditional on the $\mathbf{s}_j$, $\boldsymbol{\theta}$ and $\tau$ is normal-gamma. Specifically, we have
$$\begin{aligned}&p(\widetilde{\mathbf{B}}, \sigma^{-2}|\{\mathbf{s}_j\}_{j=1}^c, \boldsymbol{\theta}, \tau) \\ &= p(\sigma^{-2}|\{\mathbf{s}_j\}_{j=1}^c, \boldsymbol{\theta}, \tau)p(\widetilde{\mathbf{B}}|\{\mathbf{s}_j\}_{j=1}^c, \boldsymbol{\theta}, \tau, \sigma^{-2}) \\ &= p(\sigma^{-2}|\{\mathbf{s}_j\}_{j=1}^c, \boldsymbol{\theta}, \tau)\prod_{j=1}^c p(\boldsymbol{\beta}_j|\mathbf{s}_j, \boldsymbol{\theta}, \tau, \sigma^{-2}).\end{aligned}$$

The marginal distribution of $\mathbf{s}_j$ conditional on $\boldsymbol{\theta}$ and $\sigma^2$ is normal, namely,
$$p(\mathbf{s}_j|\boldsymbol{\theta}, \sigma^2) = \mathcal{N}(\mathbf{s}_j|\mathbf{0}, \sigma^2\mathbf{Q}_j), \qquad (19)$$
where $\mathbf{Q}_j = \mathbf{I}_{n_j} + \widetilde{\mathbf{K}}_j\boldsymbol{\Sigma}^{-1}\widetilde{\mathbf{K}}'_j$. We then obtain the updates of $\sigma^{-2}$ and $\boldsymbol{\beta}_j$, $j=1,\ldots,c$, as
$$\sigma^{-2}|\cdots \sim \mathcal{G}\left(\sigma^{-2}\Big|\frac{a_\sigma+nc}{2}, \frac{b_\sigma+\sum_{j=1}^c(\mathbf{s}'_j\mathbf{Q}_j^{-1}\mathbf{s}_j)}{2}\right),$$
$$\boldsymbol{\beta}_j|\cdots \sim \mathcal{N}(\boldsymbol{\beta}_j|\boldsymbol{\Psi}_j^{-1}\widetilde{\mathbf{K}}'_j\mathbf{s}_j, \sigma^2\boldsymbol{\Psi}_j^{-1}), \; j \in C,$$
where $\boldsymbol{\Psi}_j = \widetilde{\mathbf{K}}'_j\widetilde{\mathbf{K}}_j + \boldsymbol{\Sigma}$. Here and later, we use "$|\cdots$" to denote conditioning on all other variables.

Since $\tau$ is only dependent on $\mathbf{B}$ and the prior given in (17) is conjugate for $\tau$, we use the Gibbs sampler to update $\tau$ from its conditional distribution, which is given by
$$\tau|\cdots \sim \mathcal{G}\left(\tau\Big|\frac{a_\tau+nc}{2}, \frac{b_\tau + \text{tr}(\mathbf{B}'\mathbf{K}\mathbf{B})/\sigma^2}{2}\right).$$

Once samples of $\mathbf{b}_0$ and $\mathbf{B}$ have been obtained, we can calculate corresponding samples of $\mathbf{w}_0$ and $\mathbf{W}$ according to (5). Recall that in our Bayesian MSVM, $\tau/\sigma^2$ corresponds to the regularization parameter $\gamma$ in (2). This shows that we can adaptively estimate the regularization parameter as well as the regression coefficients.

In principle, we can also obtain Bayesian posterior estimates of the kernel parameter $\boldsymbol{\theta}$ using MH updates.

In practice, however, the computational complexity of these updates is likely to be prohibitive for large-scale problems. In particular, the MH sampler needs to compute the determinants of consecutive kernel matrices $\mathbf{K}$ and $\mathbf{K}^*$ in the calculation of the acceptance probability. The computational complexity of these computations can be mitigated using the Sherman-Morrison-Woodbury formula, but it remains daunting. Alternatives include setting $\boldsymbol{\theta}$ via cross-validation or via an empirical Bayes approach based on the marginal likelihood. With fixed $\boldsymbol{\theta}$, our MCMC algorithm becomes more efficient for estimating other parameters. We only need to calculate the kernel matrix once during each step. Thus, we can exploit the Sherman-Morrison-Woodbury formula for large datasets.

### 4.2 Prediction

Given a new input vector $\mathbf{x}_*$, the posterior distribution of the corresponding class label $y_*$ is given by

$$p(y_*|\mathbf{x}_*, \mathbf{y}) = \int p(y_*|\mathbf{x}_*, \Omega, \mathbf{y}) p(\Omega|\mathbf{y}) d\Omega,$$

where $\Omega$ is the vector of all model parameters. Since the above integral is analytically intractable, it is approximated via our data augmentation algorithm. Specifically, we approximate

$$p(y_* \neq j|\mathbf{x}_*, \mathbf{y}) \approx \frac{1}{T} \sum_{t=1}^{T} p\left(y_* \neq j|y_*\mathbf{x}_*, \Omega^{(t)}\right) \quad (20)$$

for $j = 1, \ldots, c$, where the $\Omega^{(t)}$ are the sampled values of $\Omega$. Finally, we allocate $\mathbf{x}_*$ to class $l$ where

$$l = \arg\min_j \{p(y_* \neq j|\mathbf{x}_*, \mathbf{y})\}.$$

## 5 Experimental Evaluation

We have conducted experiments to test the performance of our proposed Bayesian multicategory support vector machine (BMSVM), comparing to the results of Lee et al. (2004). Specifically, we test the BMSVM on four datasets in the UCI data repository: *wine*, *glass*, *waveform* and *vehicle*. Following Lee et al. (2004), in the case of the wine and glass datasets we use a leave-one-out technique to evaluate the classification results, for the waveform data, we use 300 training samples and 4,700 test samples, and for the vehicle data we use 300 training samples and 346 test samples. On the latter two datasets our results are averaged over 10 random splits.

In our experiments, all the inputs are normalized to have zero mean and unit variance. We use a Gaussian kernel with a single parameter. For the vehicle data we update the kernel parameter using Metropolis-Hastings under a product-uniform prior; for the other data sets the computational burden of using Metropolis-Hastings was prohibitive and we used cross-validation on a grid (the values were 3.5 for both the wine and waveform datasets and 10 for the glass dataset.

In our inference method, we select $M1 = 10,000$, $M2 = 5,000$, $M = 10$ and $\eta = 1000$, $a_\sigma = 3$, $b_\sigma = 10$, $a_\tau = 4$ and $b_\tau = 0.1$. When updating the kernel parameter for the vehicle data, we set the range of the uniform distributions to be $a_\theta = 0.1$ and $b_\theta = 200$. The initial values of $\sigma^2$ and $\tau$ were randomly generated from the prior distributions. For $i = 1, \ldots, n$, $z_{ij}$ is initialized as 1 if $\mathbf{x}_i$ belongs to class $j$ and as $-1/(c-1)$ otherwise.

The results are shown in Table 1, where for comparison we include the results reported in Lee et al. (2004) for the standard MSVM, the one-versus-rest binary SVM (OVR) and the alternative multi-class SVM (AltMSVM) (Guermeur, 2002). As can be seen, the classification accuracy of the Bayesian method is comparable to the accuracy of the large-margin methods.

Table 1: Test Error Rates

|         | Wine   | Glass  | Waveform | Vehicle |
|---------|--------|--------|----------|---------|
| BMSVM   | 0.0169 | 0.2383 | 0.1655   | 0.0816  |
| MSVM    | 0.0169 | 0.3645 | 0.1564   | 0.0694  |
| OVR     | 0.0169 | 0.3458 | 0.1753   | 0.0809  |
| AltMSVM | 0.0169 | 0.3170 | 0.1696   | 0.0925  |

## 6 Conclusions and Further Directions

We have presented a probabilistic formulation of the multi-class SVM and developed a Bayesian inference procedure for this architecture. Our empirical experiments have shown that the classification accuracy of the Bayesian approach is comparable to that of non-Bayesian approaches; thus, the ability of the Bayesian approach to provide estimates of uncertainty in predictions and parameter estimates does not necessarily come at a loss in classification accuracy.

Our derivation of the BMSVM in Section 2.2 involved the use of a prior distribution to cancel the normalization associated with the conditional probability of the class labels. It is also possible to consider a second approach that can be viewed as a multi-class extension of the Bayesian binary SVM due to Mallick et al. (2005). In this approach we directly evaluate the normalizing

constant. Specifically, we first rewrite (6) as:

$$p(y_i{=}j|\mathbf{f}(\mathbf{x}_i)) \propto \frac{\exp\left\{-\sum_{l=1}^{c}\left(f_l(\mathbf{x}_i)+\frac{1}{c-1}\right)_+\right\}}{\exp\left\{-\left(f_j(\mathbf{x}_i)+\frac{1}{c-1}\right)_+\right\}}$$

$$= \frac{\exp\left\{\left(f_j(\mathbf{x}_i)+\frac{1}{c-1}\right)_+\right\}}{\exp\left\{\sum_{l=1}^{c}\left(f_l(\mathbf{x}_i)+\frac{1}{c-1}\right)_+\right\}}.$$

The denominator of the above second line does not rely on $j$. Again considering that $\sum_{j=1}^{c} p(y_i{=}j|\mathbf{f}(\mathbf{x}_i)) = 1$, we define the following probabilistic model:

$$p(y_i{=}j|\mathbf{f}(\mathbf{x}_i)) = \frac{\exp\left\{\left(f_j(\mathbf{x}_i)+\frac{1}{c-1}\right)_+\right\}}{\sum_{l=1}^{c}\exp\left\{\left(f_l(\mathbf{x}_i)+\frac{1}{c-1}\right)_+\right\}} \quad (21)$$

which provides an alternative to the approach that we have pursued here.

It is interesting to consider both alternatives in the case of the binary SVM, i.e., when $c = 2$. With the approach based on cancelation from the prior, (9) reduces to

$$p(y_i{\neq}j|\mathbf{f}(\mathbf{x}_i)) \propto \exp\left\{-(f_j(\mathbf{x}_i)+1)_+\right\}, \; j = 1, 2.$$

Moreover, we have

$$p(y_i{=}2|\mathbf{f}(\mathbf{x}_i)) \propto \exp\left\{-(1+f_1(\mathbf{x}_i))_+\right\},$$
$$p(y_i{=}1|\mathbf{f}(\mathbf{x}_i)) \propto \exp\left\{-(1-f_1(\mathbf{x}_i))_+\right\}$$

due to $p(y_i{=}2|\mathbf{f}(\mathbf{x}_i)) = p(y_i{\neq}1|\mathbf{f}(\mathbf{x}_i))$, $p(y_i{=}1|\mathbf{f}(\mathbf{x}_i)) = p(y_i{\neq}2|\mathbf{f}(\mathbf{x}_i))$ and $f_1(\mathbf{x}_i) + f_2(\mathbf{x}_i) = 0$. Streamlining the equation by replacing $y = 2$ with $y = -1$ yields

$$p(y_i|f(\mathbf{x}_i)) \propto \exp\left\{-(1-y_i f(\mathbf{x}_i))_+\right\},$$

where

$$f(\mathbf{x}_i) = w_0 + \sum_{j=1}^{n} w_j K(\mathbf{x}_i, \mathbf{x}_j).$$

Let $\boldsymbol{\omega} = (w_1, \ldots, w_n)'$ and $q(\boldsymbol{\omega}) = \mathcal{N}(\boldsymbol{\omega}|\mathbf{0}, (\lambda \mathbf{K})^{-1})$. The minimization of $-\log p(\mathbf{y}, \boldsymbol{\omega})$ with respect to the $w_i$ leads us to the primal optimization problem for the binary SVM as

$$\min_{\boldsymbol{\omega}} \left\{ \sum_{i=1}^{n} \left(1 - y_i f(\mathbf{x}_i)\right)_+ + \frac{\lambda}{2} \boldsymbol{\omega}' \mathbf{K} \boldsymbol{\omega} \right\}.$$

Note that this differs from the Bayesian binary SVM of Mallick et al. (2005), which is based on the prior distribution $q(\boldsymbol{\omega}) = \mathcal{N}(\mathbf{0}, \boldsymbol{\Lambda}^{-1})$ where $\boldsymbol{\Lambda}$ is an $n{\times}n$ diagonal matrix.

On the other hand, let us consider the alternative approach based on (21). In the binary setting this yields

$$p(y_i|f(\mathbf{x}_i)) = \begin{cases} \frac{1}{1+\exp[-2y_i f(\mathbf{x}_i)]} & |f(\mathbf{x}_i)| \leq 1, \\ \frac{1}{1+\exp[-y_i(f(\mathbf{x}_i)+\mathsf{sgn}(f(\mathbf{x}_i)))]} & \text{otherwise,} \end{cases}$$

where $\mathsf{sgn}(\cdot)$ is the sign function. This model is the same as that used in the Bayesian binary SVM by Mallick et al. (2005).

It is also worth mentioning the connection to Gaussian processes. Note first that $\mathbf{KW} = \mathbf{KBH} \sim \mathcal{N}_{n,c}(\mathbf{0}, \tau^{-1}\mathbf{K}{\otimes}\mathbf{H})$, which is a singular matrix-variate distribution. Moreover, $\mathbf{Z} \sim \mathcal{N}_{n,c}(\mathbf{Z}|\mathbf{1}_n \mathbf{w}_0', \sigma^2 \mathbf{R}{\otimes}\mathbf{H})$ with $\mathbf{R} = \mathbf{I}_n + \tau^{-1}\mathbf{K}$. Whatever the value of the intercept term $\mathbf{w}_0 = \mathbf{0}$, in the multicategory SVM, we obtain $\mathbf{Z} \sim \mathcal{N}_{n,c}(\mathbf{Z}|\mathbf{0}, \sigma^2 \mathbf{R}{\otimes}\mathbf{H})$, and we see that both $\mathbf{Z}$ and $\mathbf{KW}$ follow normal distributions as they would in the Gaussian process setting. In fact, there exist interesting connections between our Bayesian MSVM and the Bayesian multinomial probit regression model of Girolami and Rogers (pear). Specifically, $\mathbf{Z}$ and $\mathbf{KW}$ correspond to latent and manifest Gaussian random matrices in the Bayesian multinomial probit regression, differing from our case in that we impose the constraints $\mathbf{Z}\mathbf{1}_c = \mathbf{0}$ and $\mathbf{KW}\mathbf{1}_c = \mathbf{0}$.

## Appendix

Let $\mathbf{w}_{\cdot j}$ and $\mathbf{b}_{\cdot j}$ be the $j$th columns of $\mathbf{W} = \mathbf{BH}$ and $\mathbf{B}$, for $j = 1, \ldots, c$. Hence, $\mathbf{w}_{\cdot j} = \mathbf{b}_{\cdot j} - \frac{1}{c}\sum_{l=1}^{c} \mathbf{b}_{\cdot l}$. Since the $\mathbf{b}_{\cdot j}$ are independent draws from $\mathcal{N}(\mathbf{0}, (\lambda \mathbf{K})^{-1})$, we have that $E(\mathbf{w}_{\cdot j}) = \mathbf{0}$ and

$$C(\mathbf{w}_{\cdot j}, \mathbf{w}_{\cdot l}) = E(\mathbf{w}_{\cdot j}\mathbf{w}_{\cdot l}') = \begin{cases} (1-c^{-1})(\lambda \mathbf{K})^{-1} & j = l, \\ -c^{-1}(\lambda \mathbf{K})^{-1} & j \neq l. \end{cases}$$

Denote $\mathsf{vec}(\mathbf{W}) = (\mathbf{w}_{\cdot 1}', \ldots, \mathbf{w}_{\cdot c}')'$. Thus, $\mathsf{vec}(\mathbf{W}) \sim \mathcal{N}(\mathbf{0}, \mathbf{H}{\otimes}(\lambda \mathbf{K})^{-1})$. It then follows from the definition of a matrix-variate normal distribution (Gupta and Nagar, 2000) that $\mathbf{W}' \sim \mathcal{N}(\mathbf{0}, \mathbf{H}{\otimes}(\lambda \mathbf{K})^{-1})$, and hence, $\mathbf{W} \sim \mathcal{N}(\mathbf{0}, (\lambda \mathbf{K})^{-1}{\otimes}\mathbf{H})$. Notice that $\mathbf{H}$ is singular because its rank is $c-1$. In fact, 0 and 1 are the eigenvalues of $\mathbf{H}$, and 1 occurs with multiplicity $c-1$. This shows that $\mathsf{vec}(\mathbf{W}')$ follows a singular matrix-variate normal distribution and its density is given by (Mardia et al., 1979):

$$\frac{(2\pi)^{-k/2}}{\prod_{i=1}^{k}\gamma_i^{1/2}} \exp\left\{-\frac{\lambda}{2}\mathsf{vec}(\mathbf{W}')' \mathbf{K}{\otimes}\mathbf{H}^{-} \mathsf{vec}(\mathbf{W}')\right\},$$

where $k$ is the rank of $(\tau \mathbf{K})^{-1}{\otimes}\mathbf{H}$, the $\lambda_i$ are the nonzero eigenvalues of $(\tau \mathbf{K})^{-1}{\otimes}\mathbf{H}$ and $\mathbf{H}^{-}$ is a generalized inverse of $\mathbf{H}$. It is easily seen that $\mathbf{H}$ itself is a generalized inverse of $\mathbf{H}$. Now suppose that $\rho_i$, $i = 1, \ldots, n$, are the eigenvalues of $\mathbf{K}$. Consider that 1 is the $(c{-}1)$-multiple eigenvalue of $\mathbf{H}$. Then the $(\lambda \rho_i)^{-1}$ are the $(c{-}1)$-multiple eigenvalues of $(\tau \mathbf{K})^{-1}{\otimes}\mathbf{H}$. This shows that

$$\prod_{i=1}^{k}\gamma_i^{1/2} = \prod_{j=1}^{n}(\lambda \rho_i)^{-(c-1)/2} = \lambda^{-n(c-1)/2}|\mathbf{K}|^{-(c-1)/2}.$$

In addition, making use of properties of Kronecker products (Muirhead, 1982, page 76), we have

$$\mathsf{vec}(\mathbf{W}')'\mathbf{K}\otimes\mathbf{H}^-\mathsf{vec}(\mathbf{W}') = \mathsf{tr}(\mathbf{K}\mathbf{W}\mathbf{W}'),$$

where we use $\mathbf{W}\mathbf{H}^- = \mathbf{W}\mathbf{H} = \mathbf{W}$. Thus, we obtain (10).

The above derivation assumes that $\mathbf{K}$ is nonsingular. When $\mathbf{K}$ is singular, a straightforward argument yields an analogous result in which $|\mathbf{K}|$ is replaced by the product of the nonzero eigenvalues of $\mathbf{K}$.